\documentclass[11pt]{article}

\usepackage[a4paper,margin=3cm]{geometry}
\usepackage{graphicx}
\usepackage{tikz}
\usepackage{pgfplots}
\pgfplotsset{compat=1.6}%{width=10cm, compat=1.9} % retrocompatibilità e proprietà varie plot
\usepgfplotslibrary{groupplots}
\usepackage{pgf}
\usetikzlibrary{positioning,shapes,arrows,fit,calc,automata,perspective,3d}
\usepackage{neuralnetwork}
\usepackage{tikz-network}
\usepackage{mathrsfs}

\usepackage[table]{xcolor}

\usepackage{algorithm}
\usepackage{hyperref}
\hypersetup{
	colorlinks=true,       % false: boxed links; true: colored links
	linkcolor=blue,%red,          % color of internal links (change box color with linkbordercolor)
	citecolor=red,%green,        % color of links to bibliography
	filecolor=magenta,      % color of file links
	urlcolor=cyan           % color of external links
}
\usepackage{cleveref}
\usepackage{booktabs}

\DeclareMathOperator{\E}{\ensuremath{\mathbb{E}}}   %   E: Expected value
\usepackage{color}
\usepackage{multirow}
\usepackage{amssymb}
\usepackage{amsmath}
\usepackage{natbib}

\newcommand{\mK}{\mathsf{K}}

\newcommand{\mD}{\mathsf{D}}

\newcommand{\mI}{\mathsf{I}}

\usepackage{yfonts} 

\DeclareMathOperator*{\argmax}{argmax}
\DeclareMathOperator*{\argmin}{argmin}

\usepackage{amsthm}
\usepackage{amsmath} 
\newtheorem{theorem}{Theorem}[section]

\newtheorem{remark}[theorem]{Remark}

\newtheorem{corollary}[theorem]{Corollary}

\newtheorem{proposition}[theorem]{Proposition}
\newtheorem{lemma}[theorem]{Lemma}

\usepackage{dsfont}
\makeatletter
 
\newenvironment{frontmatter}{}{}
 
\let\plain@oldmaketitle\maketitle
\renewcommand{\maketitle}{%
  \plain@oldmaketitle
  \begin{center}
    \small
    \plain@affiliations
    \plain@emails
  \end{center}
}
 
\let\plain@oldauthor\author
\renewcommand{\author}[2][]{%
  \ifx\@author\@empty
    \gdef\@author{#2}%
  \else
    \g@addto@macro\@author{\and #2}%
  \fi
}
 
\newcommand{\plain@affiliations}{}
\newcommand{\plain@emails}{}
 
\newcommand{\affiliation}[2][]{%
  \g@addto@macro\plain@affiliations{#2\par}%
}
 
\newcommand{\ead}[1]{%
  \g@addto@macro\plain@emails{\texttt{#1}\quad}%
}
 
\newenvironment{keyword}
  {\par\smallskip\noindent\textbf{Keywords: }\ignorespaces}
  {\par}
 
\providecommand{\sep}{\unskip;\enspace}
\providecommand{\MSC}[1][]{%
  \par\smallskip\noindent\textbf{MSC #1: }\ignorespaces
}
 
\makeatother

 \author{
L. Bruni Bruno$^{1,2}$\thanks{\texttt{ludovico.brunibruno@polito.it}}
\and
P. Massa$^{2,3}$\thanks{\texttt{paolo.massa@fhnw.ch}}
\and
E. Perracchione$^{1,2}$\thanks{\texttt{emma.perracchione@polito.it}}
\and
M. Trombini$^{1,2}$\thanks{\texttt{matteo.trombini@polito.it}}
}

\begin{document}
 
\begin{frontmatter}



% \title{SURESTOP\textgoth{4}pG: a SURE-based STOPping rule for p-Greedy kernel regression}

\title{%SURESTOP{\cal 4}$p$G: a SURE-based STOPping rule for p-Greedy kernel regression
%$p$ESTO: a $p$-grEedy kernel regression STOpping rule via risk estimation\\
{$p$ESTO: Iterative $p$-greedy kernel regression with risk-driven Early STOpping}}

\maketitle
\begin{center}
\small
$^{1}$ Dipartimento di Scienze Matematiche Giuseppe Luigi Lagrange,
Politecnico di Torino, Italy\\
$^{2}$ Gruppo Nazionale per il Calcolo Scientifico IN$\delta$AM, Rome, Italy\\
$^{3}$ Institute for Data Science, University of Applied Sciences and Arts
Northwestern Switzerland, Switzerland
\end{center}

\begin{abstract}
Kernel-based greedy algorithms select suitable approximation nodes by minimizing point-wise error indicators, such as the power function; in this context, they are usually referred to as $p$-greedy approaches. In the presence of noisy data, deterministic criteria may fail, as the discrepancy between the true solution and its estimate is influenced by both approximation error and noise amplification, and the balance between the two is unknown. This study explicitly separates these two contributions, which are governed by the power function and a Lebesgue-type term, respectively. 
These bounds motivate the use of the \textit{Stein’s unbiased risk estimator (SURE)} as a proxy of the mean squared error, which in turn provides a principled and computable stopping criterion for the number of nodes to select in the {$p$-greedy procedure}. Numerical experiments demonstrate that this approach effectively balances deterministic accuracy and statistical noise contributions.
\end{abstract}


\end{frontmatter}





\section{Introduction}

Given a set of measurements sampled at multivariate scattered points, a kernel-based greedy scheme partitions the data into a training set and a {validation set. }%test set.
The training set is used to construct an initial coarse interpolant, which is then progressively refined by iteratively adding the most informative samples selected from the 
% test set
{validation set} according to a prescribed rule. Such selection rules may rely on residuals or on deterministic error indicators, such as the power function, that exclusively depends on the kernel and on the node locations. In the interpolation setting, the latter, namely the $p$-greedy scheme, provides a reliable and well-established criterion for a priori node selection. 
% However, when observations are affected by noise, relying solely on deterministic bounds may lead to suboptimal performances.
However, when observations are affected by noise, relying solely on the power function may lead to suboptimal performances.

{Kernel regression in the presence of noise has been extensively studied (see, e.g., \cite[Chapter 15]{Fasshauer15}).
Likewise, greedy techniques for the corresponding problem have been considered \cite{santin2018greedyregularizedkernelinterpolation}.
However, to the best of our knowledge, a stopping rule for the respective greedy algorithms is still missing.
Of course, the availability of a stopping criterion is of interest not only from a theoretical perspective, but also for computational efficiency.
Typical stopping rules for kernel methods in the absence of noise can be based on the analysis of errors or residuals \cite{WenzelSantinHaasdonk}.
In the general case of iterative methods with noisy data, one can also consider the discrepancy principle introduced by Morozov \cite{morozov1966regularization}, Generalized Cross Validation (GCV) \cite{GCV}, and risk-based criteria \cite{BENVENUTO20191}.
Among these methods, the \textit{Stein’s unbiased risk estimator (SURE)} \cite{stein1981estimation} has been employed as a data-driven criterion for parameter selection \cite{raskutti2014early}.
In the present work, we propose to employ such an estimator to identify the optimal iteration of greedy methods applied to kernel interpolation in the presence of noise.
This is of interest, for instance, in combination with inverse problems for the purpose of image reconstruction \cite{bruno2025greedytechniquesinverseproblems}.}

% To address this issue{, data-driven strategies for model selection and stopping criteria have been widely investigated in the presence of noisy observations. Classical approaches include a posteriori regularization rules such as the discrepancy principle introduced by Morozov \cite{morozov1966regularization}, while risk-based approaches exploit estimates of the prediction error to select the model complexity or tuning parameters. In particular, the \textit{Stein’s unbiased risk estimator (SURE)} has been employed as a data-driven criterion for tuning parameter selection \cite{ulfarsson2013tuning}, and data-dependent early stopping has also been investigated in the context of non-parametric regression in Reproducing Kernel Hilbert Spaces (RKHSs) \cite{raskutti2014early}. In the $p$-greedy setting considered here, however, the quantity controlling the model complexity is the number of interpolation nodes selected by the greedy procedure. This quantity is of interest, for instance, in combination with inverse problems for the purpose of image reconstruction \cite{bruno2025greedytechniquesinverseproblems}.

{The goal of the present work is to study error estimates for noisy data that can be used as the target quantity of iterative greedy methods.
In particular, we show that a statistically equivalent derivation of the power-function error bound in Reproducing Kernel Hilbert Spaces (RKHSs) needs to account for the trade-off between approximation accuracy and noise amplification.}
% The present work explicitly accounts for the trade-off between approximation accuracy and noise amplification by deriving probabilistic bounds for the expected interpolation error in Reproducing Kernel Hilbert Spaces (RKHSs). 
These non-sharp bounds, involving the power function, a Lebesgue-type term, and a noise contribution, are not computable in practice.
Indeed, they require the knowledge of the norm of exact function, while only noisy observations are available.
% This motivates the use of the 
% Stein's Unbiased Risk Estimator (SURE) \cite{stein1981estimation} 
{This motivates the use of SURE in this context}
as a proxy for the Mean Squared Error (MSE). 
Building upon this characterization, {the $p$ESTO algorithm, proposed in this paper,} provides a principled stopping rule for the iterative greedy selection by balancing approximation error and statistical noise effects.

The problem we deal with reads as follows, and we refer the reader to the monographs \cite{Fasshauer15,De_Marchi_Marchetti_Perracchione_2026,Wendland_2004} for further details on kernel-based methods. 
Let $\kappa \colon \Omega \times \Omega \to \mathbb{R}$ be a strictly positive definite kernel and $\mathcal{H}_\kappa$ the corresponding RKHS. Let $X_n = \{\boldsymbol{x}_1, \dots, \boldsymbol{x}_n \} \subseteq \Omega \subseteq \mathbb{R}^d$ be a set of nodes and let $f \in \mathcal{H}_\kappa(\Omega)$ be the unknown function to approximate
{for which noisy observations are available}. 
We assume that the associated observations are realizations of a Gaussian random variable $\boldsymbol{y}_n$ {with independent components}, namely $y_i \coloneqq 
(\boldsymbol{y}_n)_i = f(\boldsymbol{x}_i) + \varepsilon_i$, $i = 1,\ldots,n,$ 
% where $
% \boldsymbol{\varepsilon}_n = (\varepsilon_1,\ldots,\varepsilon_n) \sim \mathcal{N}(\textbf{0}, \sigma_{n}^2 \mI_n). $
{where $
\boldsymbol{\varepsilon}_n = (\varepsilon_1,\ldots,\varepsilon_n) \sim \mathcal{N}(\textbf{0}, {\rm diag}(\sigma^2_1, \dots, \sigma^2_n)). $}
Consequently, 
% $\boldsymbol{y}_n = (y_1, \ldots, y_n) \sim \mathcal{N}(\boldsymbol{f}_n, \sigma_n^2 \mI_n),$
% where $\boldsymbol{f}_n = (f_1, \ldots, f_n)^{\intercal} = (f(\boldsymbol{x}_1), \ldots, f(\boldsymbol{x}_n))^{\intercal}$.
{$\boldsymbol{y}_n = (y_1, \ldots, y_n) \sim \mathcal{N}(\boldsymbol{f}_n, {\rm diag}(\sigma^2_1, \dots, \sigma^2_n)),$}
where $\boldsymbol{f}_n = (f_1, \ldots, f_n)^{\intercal} = (f(\boldsymbol{x}_1), \ldots, f(\boldsymbol{x}_n))^{\intercal}$.
With abuse of notation, we denote both the random variables and their realizations by the same boldface symbol.
% We further assume that $\sigma^2_n$ is constant for each $n$, i.e., $\sigma_n^2 = \sigma^2$. 
{We further assume that the noise variance is constant, i.e., $\sigma^2_i = \sigma^2$ for $i=1,\dots,n$. }

In RKHSs, the problem consists in computing the regularized interpolant
\begin{equation}\label{eq:interpolant}
\min_{{s}\in{\cal H}_{\kappa}}\;
\sum_{i=1}^n (y_i - {s}(\boldsymbol{x}_i))^2
+ \lambda \|{s}\|_{{\cal H}_{\kappa}}^2, \quad \lambda \in [0,+\infty),
\end{equation}
whose solution is
$$
s_{X_n,\boldsymbol{y}_n}(\boldsymbol{x})
= \sum_{i=1}^n c_i\, \kappa(\boldsymbol{x},\boldsymbol{x}_i),
$$
where $\boldsymbol{c}_n = (c_1,\ldots,c_n)^{\intercal} = (\mK_n+\lambda \mI_n)^{-1}\boldsymbol{y}_n$, $(\mK_n)_{ij} \coloneqq \kappa(\boldsymbol{x}_i,\boldsymbol{x}_j)$,  $i,j = 1,\ldots,n$, and $\mI_n$ is the $n\times n$ identity matrix. 
% {When the matrix $\mK_n$ is positive definite, for any $ \lambda \geq 0 $ the existence and uniqueness of the regularized interpolant $ s_{X_n, \boldsymbol{y}_n} $ follows from the generalized representer theorem \cite{GenReprThm};
% in the present paper, we will consider kernels to whom we can associate positive definite matrices.
% Otherwise, if $\mK_n$ is only positive semidefinite, we refer the reader to \cite[Theorem 1]{santin2018greedyregularizedkernelinterpolation}.}
{Under the assumptions listed above, $\mK_n$ is positive definite and the existence and uniqueness of the regularized interpolant $ s_{X_n, \boldsymbol{y}_n} $ is guaranteed by the generalized representer theorem \cite{GenReprThm} for any $ \lambda \geq 0 $. 
In the case when $\mK_n$ is only positive semidefinite, the uniqueness of the interpolant coefficients is no longer guaranteed for $\lambda=0$ \cite[Theorem 1]{santin2018greedyregularizedkernelinterpolation}.}

% Greedy methods construct surrogate models $ s $ for $ f $ by selecting a subset of the data available at $X_n$.
Greedy methods construct surrogate models for $ f $ by selecting a subset of the data available at $X_n$.
At the $(j+1)$-th step of the iteration, a new node is selected as $
\boldsymbol{x}_{j+1} = \argmax_{\boldsymbol{x} \in X_n \setminus X_{j}} \eta(\boldsymbol{x}),$ where $X_{j} \subseteq X_n$
% At the $j$-th step of the iteration, a new node is selected as $
% \boldsymbol{x}_j = \argmax_{\boldsymbol{x} \in X_n \setminus X_j} \eta(\boldsymbol{x}),$ 
% where $X_j \subseteq X_n$ 
and $\eta$ is a given error indicator, such as the power function; we refer the reader to \cite{DeMarchi2005317,Wenzel2021,WenzelSantinHaasdonk}. 
% However, the latter yields a purely deterministic bound that neglects noise, thereby motivating the use of regularization as in \eqref{eq:interpolant}.  
The objective of this work is to develop a new error bound that balances approximation accuracy and noise amplification.
Furthermore, as this bound is not computable in practice, we utilize the SURE capturing both contributions. 
The resulting method, referred to as 
$p$ESTO, provides a reliable estimate of the MSE and yields an a priori stopping criterion for the $p$-greedy scheme \cite{bruno2025greedytechniquesinverseproblems}.
% The objective of this work is to develop a hybrid estimator that captures both deterministic and stochastic contributions. Since the method provides a reliable estimate of the MSE, the resulting method, referred to as 
% $p$ESTO, yields as a benefit an a priori stopping criterion for the $p$-greedy scheme \cite{bruno2025greedytechniquesinverseproblems}.

The paper is organized as follows. Section \ref{sec:interpolation_error} presents the expected approximation error. Section \ref{sure} highlights the connection with SURE. Section \ref{sec:numerical_experiments} reports numerical results supporting our findings, and conclusions are drawn in Section \ref{Conclusion}.

\section{Expected interpolation error}
\label{sec:interpolation_error}
{To begin with, we analyze the expected value of the error in the presence of noisy data. 
The following results hold for any subset $X_j$ of the set of interpolation nodes $X_n$. 
In the subsequent sections, we will in fact propose a greedy strategy for selecting such a subset.}
% We focus on the $j$-th step of the greedy procedure, where one only considers the set of nodes $X_j$ and the corresponding data $\boldsymbol{y}_j$ to compute the interpolant. 
{The results presented in this section provide theoretical error estimates which may be hard to compute in practice.
For this reason, in the next section we will introduce a {statistically-based} and computable proxy.}

Letting $\boldsymbol{k}_j(\boldsymbol{x})^\intercal = (\kappa(\boldsymbol{x},\boldsymbol{x}_1),\ldots,\kappa(\boldsymbol{x},\boldsymbol{x}_j))$ and 
$\boldsymbol{c}_j = (\mK_j+\lambda \mI_j)^{-1} \boldsymbol{y}_j$, we obtain
\begin{align*}
s_{X_j,\boldsymbol{y}_j}(\boldsymbol{x}) & = \sum_{i=1}^j c_i \, \kappa(\boldsymbol{x},\boldsymbol{x}_i) = \boldsymbol{k}_j(\boldsymbol{x})^\intercal (\mK_j+\lambda \mI_j)^{-1} \boldsymbol{y}_j = \boldsymbol{\ell}_{\lambda,j} (\boldsymbol{x})^{\intercal} \boldsymbol{y}_j \\
& = \boldsymbol{\ell}_{\lambda,j} (\boldsymbol{x})^{\intercal} \boldsymbol{f}_j+\boldsymbol{\ell}_{\lambda,j} (\boldsymbol{x})^{\intercal} \boldsymbol{\varepsilon}_j= s_{X_j,\boldsymbol{f}_j}(\boldsymbol{x})+s_{X_j,\boldsymbol{\varepsilon_j}}(\boldsymbol{x}) ~,
\end{align*} 
where $\boldsymbol{\ell}_{\lambda,j} (\boldsymbol{x})^{\intercal} = \boldsymbol{k}_j(\boldsymbol{x})^\intercal (\mK_j+\lambda \mI_j)^{-1}$ are known as the regularized Lagrange or cardinal functions.
% The following computations extend the well-known results in the noise-free setting; we refer the reader to, e.g., \cite[\S 9]{Fasshauer15}.
{The following proof follows the classical arguments used in the noise-free setting (see, e.g., \cite[\S 9]{Fasshauer15}), while incorporating the additional contribution due to the observation noise.}

\begin{proposition}\label{prop:exp_err_bound}
{For any $f \in \mathcal{H}_k$ and $\boldsymbol{x} \in\Omega$, the} expected value of the squared interpolation error satisfies the following upper bound

\begin{align}\label{eq:expected_error_bound}
\mathbb{E}_{\boldsymbol{y}_n} \left[ \left\vert f(\boldsymbol{x}) - s_{X_j,\boldsymbol{y}_j}(\boldsymbol{x}) \right\vert^2 \right] & = \mathbb{E}_{\boldsymbol{y}_j} \left[ \left\vert f(\boldsymbol{x}) - s_{X_j,\boldsymbol{y}_j}(\boldsymbol{x}) \right\vert^2 \right]  \nonumber
\\
& \leq 2\Vert f \Vert_{\mathcal{H}_\kappa}^2 P^2_{X_j,\kappa}(\boldsymbol{x}) + 2j \sigma^2 \Lambda_{X_j,\kappa,{\lambda}}^2(\boldsymbol{x})~, 
\end{align}
where
\begin{align}\label{eq:power_f}
P_{X_j,\kappa}^2(\boldsymbol{x}) \coloneqq  \kappa(\boldsymbol{x},\boldsymbol{x})
& -2\,\boldsymbol{k}_j(\boldsymbol{x})^\intercal (\mK_j + \lambda \mI_j)^{-1}\boldsymbol{k}_j(\boldsymbol{x}) \\
& +\boldsymbol{k}_j(\boldsymbol{x})^\intercal (\mK_j + \lambda \mI_j)^{-1} \mK_j (\mK_j+\lambda \mI_j)^{-1}\boldsymbol{k}_j(\boldsymbol{x}) \nonumber
\end{align}
is the squared power function introduced in \cite[Equation (25)]{santin2018greedyregularizedkernelinterpolation} and
\begin{equation*}
\Lambda_{X_j,\kappa,{\lambda}}^2(\boldsymbol{x}) \coloneqq \sum_{h=1}^j (\boldsymbol{\ell}_{\lambda,j} (\boldsymbol{x}))_h^2 
\end{equation*}
is the sum of the squared regularized Lagrange functions {defined in \cite[Equation (22)]{santin2018greedyregularizedkernelinterpolation} and referred to as \textquotedblleft $ \ell_2$-Lebesgue function\textquotedblright}.
\end{proposition}

\begin{proof}
Since $s_{X_j,\boldsymbol{f}_j}(\boldsymbol{x}) = \boldsymbol{\ell}_{\lambda,j} (\boldsymbol{x})^{\intercal} \boldsymbol{f}_j$, by \cite[Equation (9.6)]{Fasshauer15} one has
\begin{equation*}
s_{X_j,\boldsymbol{f}_j}(\boldsymbol{x}) = \boldsymbol{k}_j(\boldsymbol{x})^\intercal (\mK_j+\lambda \mI_j)^{-1} \boldsymbol{f}_j = \langle {f}, \boldsymbol{k}_j(\cdot)^\intercal (\mK_j +\lambda \mI_j)^{-1} \boldsymbol{k}_j(\boldsymbol{x}) \rangle_{{\cal H}_\kappa} ~.
\end{equation*}
Hence, by the reproducing property and the Cauchy-Schwarz inequality
\begin{align*}
\left\vert f(\boldsymbol{x}) - s_{X_j,\boldsymbol{f}_j} (\boldsymbol{x})\right\vert & = \left\vert \langle f, \kappa(\cdot, \boldsymbol{x}) \rangle - \langle f, \boldsymbol{k}_j(\cdot)^\intercal (\mK_j +\lambda \mI_j)^{-1} \boldsymbol{k}_j(\boldsymbol{x}) \rangle_{{\cal H}_\kappa}\right\vert \\
& \leq \Vert f \Vert_{\mathcal{H}_\kappa} \Vert \kappa(\cdot, \boldsymbol{x}) -  \boldsymbol{k}_j(\cdot)^\intercal (\mK_j +\lambda \mI_j)^{-1} \boldsymbol{k}_j(\boldsymbol{x}) \Vert_{\mathcal{H}_\kappa} ~.
\end{align*}
In particular, recalling that the power function $P_{X_j,\kappa}(\boldsymbol{x})$ satisfies 
\begin{equation*}
\begin{split}
P^2_{X_j,\kappa}(\boldsymbol{x}) &\coloneqq \Vert \kappa(\cdot, \boldsymbol{x}) -  \boldsymbol{k}_j(\cdot)^\intercal (\mK_j+\lambda \mI_j)^{-1} \boldsymbol{k}_j(\boldsymbol{x}) \Vert_{\mathcal{H}_\kappa}^2 \\
& =\kappa(\boldsymbol{x},\boldsymbol{x})
-2\,\boldsymbol{k}_j(\boldsymbol{x})^\intercal (\mK_j + \lambda \mI_j)^{-1}\boldsymbol{k}_j(\boldsymbol{x}) \\ & \;\quad +\boldsymbol{k}_j(\boldsymbol{x})^\intercal (\mK_j + \lambda \mI_j)^{-1} \mK_j (\mK_j+\lambda \mI_j)^{-1}\boldsymbol{k}_j(\boldsymbol{x}) ~,
\end{split}
\end{equation*}
we have $\left\vert f(\boldsymbol{x}) - s_{X_j,\boldsymbol{f}_j} (\boldsymbol{x})\right\vert \leq \Vert f \Vert_{\mathcal{H}_\kappa} P_{X_j,\kappa}^2(\boldsymbol{x})$.
Moreover, by the Cauchy-Schwarz inequality and basic properties of normal distributions
\begin{align*}
\mathbb{E}_{\boldsymbol{\varepsilon}_j} \left[ \left\vert s_{X_j,\varepsilon_j}(\boldsymbol{x})\right\vert^2 \right] & =
\mathbb{E}_{\boldsymbol{\varepsilon}_j} \left[ \left\vert \boldsymbol{k}_j(\boldsymbol{x})^\intercal (\mK_j + \lambda \mI_j)^{-1} \boldsymbol{\varepsilon}_j \right\vert^2 \right] \\
& \leq \mathbb{E}_{\boldsymbol{\varepsilon}_j} \left[ \sum_{h=1}^j (\boldsymbol{\varepsilon}_j)_h^2 \right] \cdot \left( \sum_{h=1}^j  (\boldsymbol{\ell}_{\lambda,j} (\boldsymbol{x}))_h^2 \right) = j \sigma^2 \Lambda_{X_j,\kappa,{\lambda}}^2(\boldsymbol{x}) ~.
\end{align*}

Since $(a+b)^2\leq 2 (a^2 + b^2)$ for any $a,b\in\mathbb{R}$, and by applying the triangular inequality, we get the bound in  \eqref{eq:expected_error_bound}, i.e.,
\begin{align}\label{eq:expected_error_bound_proof}
\mathbb{E}_{\boldsymbol{y}_j} \left[ \left\vert f(\boldsymbol{x}) - s_{X_j,\boldsymbol{y}_j}(\boldsymbol{x}) \right\vert^2 \right] & \leq 2 \left\vert f(\boldsymbol{x}) - s_{X_j,\boldsymbol{f}_j}(\boldsymbol{x}) \right\vert^2 + 2 \, \mathbb{E}_{\boldsymbol{\varepsilon}_j} \left[ \left\vert s_{X_j,\varepsilon_j}(\boldsymbol{x})\right\vert^2 \right] \nonumber \\ & \leq 2\Vert f \Vert_{\mathcal{H}_\kappa}^2 P^2_{X_j,\kappa}(\boldsymbol{x}) + 2j \sigma^2 \Lambda_{X_j,\kappa,{\lambda}}^2(\boldsymbol{x}) ~.
\end{align}
The above obtained \eqref{eq:expected_error_bound_proof} gives the inequality in \eqref {eq:expected_error_bound}. The remaining equality in \eqref {eq:expected_error_bound} immediately follows observing that $\left\vert f(\boldsymbol{x}) - s_{X_j,\boldsymbol{y}_j}(\boldsymbol{x}) \right\vert^2$ does not depend on the elements in $\boldsymbol{y}_n \setminus \boldsymbol{y}_j$.
\end{proof}

% The right-hand side of Equation \eqref{eq:expected_error_bound} contains two terms.
% The first one accounts for the usual approximation error, while the second encodes information on the amplification of noise.
% The balance between the two terms is given by the signal-to-noise ratio $\Vert f \Vert_{\mathcal{H}_\kappa} / \sigma$, which is unknown in practice.
% Hence, classical $p$-greedy approaches, which only hinge on the minimization of the power function, cannot be directly applied.

\begin{remark}
    Let $ \mathscr{L}_{X_j,\kappa,{\lambda}} (\boldsymbol{x}) \coloneqq \sum_{h=1}^j |(\boldsymbol{\ell}_{\lambda,j} (\boldsymbol{x}))_h | $ be the usual Lebesgue function. Then, since its terms are positive, one clearly has
    $$ \Lambda_{X_j,\kappa,{\lambda}}^2 (\boldsymbol{x}) = \sum_{h=1}^j (\boldsymbol{\ell}_{\lambda,j} (\boldsymbol{x}))_h^2 \leq \left( \sum_{h=1}^j |(\boldsymbol{\ell}_{\lambda,j} (\boldsymbol{x}))_h| \right)^2 = \mathscr{L}_{X_j,\kappa,{\lambda}}^2 (\boldsymbol{x}) .$$
    That is, the square of the Lebesgue constant $ \sup_x \mathscr{L}_{X_j,\kappa,{\lambda}} (\boldsymbol{x})$ is a (non sharp) upper bound for $ \Lambda_{X_j,\kappa,{\lambda}}^2 (\boldsymbol{x}) $. The role of the Lebesgue constant in kernel approximation is of independent interest \cite{Themistoclakis}.
\end{remark}

\begin{remark}
     {In \cite{santin2018greedyregularizedkernelinterpolation}, the term $ \Lambda_{X_j, \kappa, \lambda} $ is related 
     % The term $ \Lambda_{X_j, \kappa, \lambda} $ derived in the above computation coincides with the \textquotedblleft $ \ell_2$-Lebesgue function\textquotedblright\ introduced in \cite[Equation (22)]{santin2018greedyregularizedkernelinterpolation}, where such a quantity is related 
     with the fill distance and the regularizing parameter $ \lambda $. In particular, in such a paper it is proven %in \cite{santin2018greedyregularizedkernelinterpolation} 
     that $ \Lambda_{X_j, \kappa, \lambda} $ is uniformly bounded by $ F_\kappa(h_{X_n}) / \sqrt{\lambda} $, being $ F_\kappa $ a positive function (depending on the kernel $ \kappa $) and $ h_{X_n} $ the \emph{fill distance} associated with the set $ X_n $. This makes the last term of \eqref{eq:expected_error_bound_proof} uniformly bounded for $ \lambda $ large enough, and %while making it 
     large for small $ \lambda $.}
\end{remark}

The right-hand side of Equation \eqref{eq:expected_error_bound} contains two terms.
The first one accounts for the usual approximation error, while the second encodes information on the amplification of noise.
The balance between the two terms is given by the signal-to-noise ratio $\Vert f \Vert_{\mathcal{H}_\kappa} / \sigma$. {The main practical limitation is that the RKHS norm $\Vert f \Vert_{\mathcal{H}_\kappa}$ is unknown and cannot be directly estimated
from the available noisy observations.
Moreover, estimating this norm is already a challenging task even in the noise-free
setting (e.g. \cite{kim2024enhancinggaussianprocesssurrogates,wenzel2026reliablesamplingbasedrkhsnorm}).}
% , which is unknown in practice as we only know  noisy observations of $f$.
Therefore, greedy approaches based on the minimization of \eqref{eq:expected_error_bound} cannot be applied in practice.
To overcome this issue, we propose $p$ESTO: we iteratively select the nodes by minimizing the power function \eqref{eq:power_f} until a probabilistic stopping criterion that takes into account noise amplification is met.

In the next Section we propose an approach levering on statistical risk to estimate the expected interpolation error on the entire set $X_n$ of the evaluation nodes.
To this end, we shall need an immediate consequence of Proposition \ref{prop:exp_err_bound}.

\begin{corollary} \label{cor:Expected}
Let $X_n = \{\boldsymbol{x}_1, \ldots, \boldsymbol{x}_n \}$ be the set of interpolation nodes.
One has
\begin{align}\label{eq:expected_error_bound_nodes}
%& \mathbb{E}_{\boldsymbol{y}_n} \left[ \frac{1}{n} \sum_{i=1}^n \left\vert f(\boldsymbol{x}_i) - s_{X_j,\boldsymbol{y}_j}(\boldsymbol{x}_i) \right\vert^2 \right] 
% \leq \nonumber \\
& \mathbb{E}_{\boldsymbol{y}_n} \left[ {{\rm MSE}\left(s_{X_j,\boldsymbol{y}_j}, \boldsymbol{f}_n \right)} \right] 
\leq \nonumber \\ 
& \leq \frac{2}{n}\Vert f \Vert_{\mathcal{H}_K}^2 \left(\sum_{i=1}^n P^2_{X_j,\kappa}(\boldsymbol{x}_i)\right) + \frac{2j}{n} \sigma^2 \left(\sum_{i=1}^n \Lambda_{X_j,\kappa,{\lambda}}^2(\boldsymbol{x}_i) \right) ~,
\end{align}
{where the Mean Squared Error (MSE) is defined as
$$
{\rm MSE}\left(s_{X_j,\boldsymbol{y}_j}, \boldsymbol{f}_n \right) \coloneqq \frac{1}{n} \sum_{i=1}^n
\left(s_{X_j,\boldsymbol{y}_j}(x_i) - f(x_i)  \right)^2 ~.
$$}
For further reference, we term $\mathcal{E}(f,\kappa, \sigma,{\lambda}, j) \coloneqq \mathcal{E}_{{P}}(f,\kappa, j) + \mathcal{E}_{{\Lambda}}(\kappa, \sigma,{\lambda}, j)$ the right-hand side of \eqref{eq:expected_error_bound_nodes}.
\end{corollary}

% {We also remark that the quantity on the left-hand side of \eqref{eq:expected_error_bound_nodes} coincides with the expected mean squared error associated with the regularized interpolant. This perspective will be further developed in the next section.}

% {In the next section we discuss a computable estimate for the left-hand side of  \eqref{eq:expected_error_bound_nodes}.}

\section{Risk estimator}
\label{sure}

{In the noisy setting, evaluations of the interpolant on the set of nodes, i.e., \\$(s_{X_j,\boldsymbol{y}_j}(\boldsymbol{x}_1), \dots, s_{X_j,\boldsymbol{y}_j}(\boldsymbol{x}_n))^{\intercal}$, depend on the random variable $\boldsymbol{y}_j$ and, therefore, form a random vector. 
Realizations of this random vector provide estimates of the unknown exact evaluations $\boldsymbol{f}_n = (f_1, \ldots, f_n)^{\intercal}$.
For this reason, such a vector can be thought of as an \emph{estimator} $\hat{f}(\boldsymbol{y}_j)$ of $\boldsymbol{f}_n$ \citep[pag. 286]{MR2208349}. }

{The estimator $(s_{X_j,\boldsymbol{y}_j}(\boldsymbol{x}_1), \dots, s_{X_j,\boldsymbol{y}_j}(\boldsymbol{x}_n))^{\intercal}$ depends on the number $j$ of interpolation nodes. 
In practice, it may be convenient to construct the interpolant over a set of nodes of cardinality $j < n$. 
This subset can be selected by a greedy procedure. 
To determine the optimal number $j$ of nodes we propose the use of the \textit{Stein’s unbiased risk estimator (SURE)} to estimate the (expected) distance of $(s_{X_j,\boldsymbol{y}_j}(\boldsymbol{x}_1), \dots, s_{X_j,\boldsymbol{y}_j}(\boldsymbol{x}_n))^{\intercal}$ from $\boldsymbol{f}_n$ and terminate the greedy procedure at the iteration $j$ minimizing such a distance. 
In contrast with the theoretical error bounds derived in Section~\ref{sec:interpolation_error}, SURE is explicitly computable since it depends only on the available noisy data. 
% SURE is termed a \emph{risk estimator} as it estimates the (average) squared $\ell^2$ distance between the unknown exact evaluations $\boldsymbol{f}_n$ and $(s_{X_j,\boldsymbol{y}_j}(\boldsymbol{x}_1), \dots, s_{X_j,\boldsymbol{y}_j}(\boldsymbol{x}_n))^{\intercal}$.
% In fact, SURE allows estimating the (average) squared $\ell^2$ distance between the unknown exact evaluations $\boldsymbol{f}_n$ and $(s_{X_j,\boldsymbol{y}_j}(\boldsymbol{x}_1), \dots, s_{X_j,\boldsymbol{y}_j}(\boldsymbol{x}_n))^{\intercal}$.
}

% {
% In the noisy setting, the objective is no longer to recover the observations themselves, but rather the underlying vector of noise-free values $f_n$. 
% Even when all the available observations are employed (i.e. $i=n$) the evaluations of the regularized interpolant \eqref{eq:interpolant} over the set of nodes $X_n$ $(s_{X_j,\boldsymbol{y}_j}(\boldsymbol{x}_1), \dots, s_{X_j,\boldsymbol{y}_j}(\boldsymbol{x}_n))$ need not coincide with the noise-free values $\boldsymbol{f}_n = (f_1, \dots, f_n)$.
% Hence, the array of evaluations $(s_{X_j,\boldsymbol{y}_j}(\boldsymbol{x}_1), \dots, s_{X_j,\boldsymbol{y}_j}(\boldsymbol{x}_n))$ can only be considered as an estimator of $(f_1, \dots, f_n)$.}

Given an estimator $\hat{f}(\boldsymbol{y}_j)$ of $\boldsymbol{f}_n$, SURE is defined as
\begin{equation} \label{eq:defSURE}
{\rm SURE}\left(\hat{f},\boldsymbol{y}_n \right) \coloneqq \frac{1}{n} \sum_{i=1}^n \left(\hat{f}_i(\boldsymbol{y}_j) - y_i  \right)^2 + \frac{2}{n} \sigma^2 \sum_{i=1}^n \partial_i \hat{f}_i(\boldsymbol{y}_j) - \sigma^2 ~,
\end{equation}
where, throughout the paper, $\partial_i$ indicates the partial derivative with respect to the $i$-th component of $\boldsymbol{y}_j$.
{SURE provides an estimate of the expected MSE distance between the unknown exact evaluations $\boldsymbol{f}_n$ and $(s_{X_j,\boldsymbol{y}_j}(\boldsymbol{x}_1), \dots, s_{X_j,\boldsymbol{y}_j}(\boldsymbol{x}_n))^{\intercal}$.
In the statistical setting, this expectation is called the \emph{risk} of $(s_{X_j,\boldsymbol{y}_j}(\boldsymbol{x}_1), \dots, s_{X_j,\boldsymbol{y}_j}(\boldsymbol{x}_n))^{\intercal}$ with respect to the MSE loss.}

SURE is an unbiased estimator of the expected value of {the MSE,}
% Mean Squared Error (MSE):
\begin{equation}\label{eq:sure_prop}
\E_{\boldsymbol{y}_n} \left[ {\rm SURE}\left(\hat{f},\boldsymbol{y}_n \right) \right] = 
\E_{\boldsymbol{y}_n}
\left[{\rm MSE}\left(\hat{f}, \boldsymbol{f}_n \right) \right] ~.
\end{equation}
% where
% $$
% {\rm MSE}\left(\hat{f}, \boldsymbol{f}_n \right) \coloneqq \frac{1}{n} \sum_{i=1}^n
% \left(\hat{f}_i(\boldsymbol{y}_j) - {f(x_i)}  \right)^2 ~.
% $$
% In the presence of noise, where the regularized interpolant \eqref{eq:interpolant} is evaluated over the set of nodes $X_n$, the evaluations $(s_{X_j,\boldsymbol{y}_j}(\boldsymbol{x}_1), \dots, s_{X_j,\boldsymbol{y}_j}(\boldsymbol{x}_n))$ need not coincide with the noise-free values $\boldsymbol{f}_n = (f_1, \dots, f_n)$ even when $j=n$.
% Hence, the array of evaluations $(s_{X_j,\boldsymbol{y}_j}(\boldsymbol{x}_1), \dots, s_{X_j,\boldsymbol{y}_j}(\boldsymbol{x}_n))$ can only be considered as an estimator of $(f_1, \dots, f_n)$.
SURE can therefore be computed to estimate the squared distance between the values of the interpolant when evaluated over the set of nodes $X_n$ and the noise-free data $\boldsymbol{f}_n$ (i.e., the left-hand side of Equation~\eqref{eq:expected_error_bound_nodes}){, provided that the noise variance $\sigma^2$ is available.
This is a standard assumption as in many practical applications the noise level is in fact either characterized by the acquisition process or estimated from the observations \cite{Benvenuto_2020}. The following holds true.}

% Specifically, the following holds true.

\begin{lemma}\label{lemma:sure}
In our setting, SURE reads as
\begin{equation} \label{eq:sureGREEDY}
{\rm SURE}(s_{X_j, \boldsymbol{y}_j}, \boldsymbol{y}_n) \coloneqq \frac{1}{n}\sum_{i=1}^n (s_{X_j, \boldsymbol{y}_j} (\boldsymbol{x}_i) - y_i)^2 + \frac{2}{n}\sigma^2 {\rm trace} \left( \mK_j (\mK_j + \lambda \mI_j)^{-1} \right) - \sigma^2 
\end{equation}
and satisfies
\begin{equation}\label{eq:sure_property}
\E_{\boldsymbol{y}_n} \left[ {\rm MSE}(s_{X_j, \boldsymbol{y}_j}, \boldsymbol{f}_n) \right]
= \E_{\boldsymbol{y}_n} \left[ {\rm SURE}(s_{X_j, \boldsymbol{y}_j}, \boldsymbol{y}_n) \right] ~.
\end{equation}
\end{lemma}

\begin{proof}
From
$$
s_{X_j,\boldsymbol{y}_j}(\boldsymbol{x}_i) = \boldsymbol{k}_j(\boldsymbol{x_i})^\intercal (\mK_j+\lambda \mI_j)^{-1} \boldsymbol{y}_j = 
\sum_{l=1}^j \left[ \mK_j \left(\mK_j + \lambda \mI_j \right)^{-1} \right]_{il} \left(\boldsymbol{y}_j\right)_l ~,
$$
it follows
$$
\partial_i s_{X_j,\boldsymbol{y}_j}(\boldsymbol{x}_i) =
\begin{cases}
\left[ \mK_j \left(\mK_j + \lambda \mI_j \right)^{-1} \right]_{ii} & \text{if $\boldsymbol{x}_i \in X_j$} \\
0 & \text{if $\boldsymbol{x}_i \in X_n \setminus X_j$}
\end{cases}
~~.
$$
Plugging this into \eqref{eq:defSURE} one finds \eqref{eq:sureGREEDY}. 
Finally, Equation \eqref{eq:sure_property} follows from \eqref{eq:sure_prop}.
\end{proof}

{\begin{remark}\label{rmk:bound_trace}
    The central term of \eqref{eq:sureGREEDY}, which represents the complexity or degrees of freedom of a statistical model \cite[p. 3146]{Deledalle_2017}, can be bounded from above as follows. Since $ \mK_j $ is symmetric, there exists an orthogonal matrix $ Q $ such that $ \mK_j = Q \mD_j Q^\top $, where $ \mD_j $ is a diagonal matrix generated by the eigenvalues of $ K_j $. Then, since $ Q^\top = Q^{-1} $ and hence $ Q Q^\top = \mI_j $, one has
    \begin{align*} \mK_j (\mK_j + \lambda \mI_j)^{-1} & = Q \mD_j Q^\top (Q \mD_j Q^\top + \lambda Q Q^\top)^{-1} = Q \mD_j Q^\top (Q (\mD_j + \lambda \mI_j) Q^\top)^{-1} \\
    & = Q \mD_j Q^\top Q^{-\top} (\mD_j + \lambda\mI_j)^{-1} Q^{-1} = Q \mD_j  (\mD_j + \lambda\mI_j)^{-1} Q^{-1} .
    \end{align*}
    Using the cyclic property of the trace, one gets
    \begin{align*} {\rm trace} \left( \mK_j (\mK_j + \lambda \mI_j)^{-1} \right) & = {\rm trace} \left( Q \mD_j  (\mD_j + \lambda\mI_j)^{-1} Q^{-1} \right) \nonumber \\
    & = {\rm trace} \left( Q^{-1}  Q \mD_j  (\mD_j + \lambda\mI_j)^{-1} \right) \nonumber \\
    & = {\rm trace} \left(\mD_j  (\mD_j + \lambda\mI_j)^{-1} \right) = \sum_{i=1}^j \frac{\mu_i}{\mu_i + \lambda }, 
    \end{align*}
    being $ \mu_i $ the $i$-th eigenvalue of $ \mD_j $ (and hence of $ \mK_j $). As, further, $ \mK_j $ is positive definite and $ \lambda > 0 $ one has $ \mu_i / (\mu_i + \lambda) \leq 1 $ for each $ i $, and this yields the chain of inequalities
    \begin{equation} \label{eq:simplificattiontrace}
    \frac{2}{n}\sigma^2 {\rm trace} \left( \mK_j (\mK_j + \lambda \mI_j)^{-1} \right) \leq 2\sigma^2 \frac{j}{n} \leq 2 \sigma^2 .
    \end{equation}
    Numerical verifications show that, if the magnitude of $ \lambda $ is smaller than the majority of the eigenvalues of $ \mK_j $, the first bound in \eqref{eq:simplificattiontrace} gives an acceptable and sensibly cheaper approximation for the central term of SURE, while the second appears to be generally too rough.
\end{remark}}

\begin{remark}
There is an analogy between $\mathcal{E}(f,\kappa, \sigma, {\lambda}, j)$ and SURE.
Indeed, the term involving the squared power function in \eqref{eq:expected_error_bound_nodes} and the average of the squared residuals $\frac{1}{n}\sum_{i=1}^n (s_{X_j, \boldsymbol{y}_j} (\boldsymbol{x}_i) - y_i)^2$ represent the approximation error and decrease for increasing number of interpolation nodes $j$.
Differently, the term involving the sum of the squared regularized Lagrange functions in \eqref{eq:expected_error_bound_nodes} and $2\sigma^2 {\rm trace} \left( \mK_j (\mK_j + \lambda \mI_j)^{-1} \right)$ describe the noise amplification and increase over the iterations of the greedy approach.
{This heuristic interpretation of the constituting terms of both the expected interpolation error and SURE suggests that the greedy approach may semi-converge, in the sense that the quality of the interpolant improves up to some iteration; however, due to noise propagation, the interpolant diverges from the ground truth function after more interpolation nodes are added.
Consistently, both the expected interpolation error and SURE reach their respective minimum at a number of iterations smaller than the total number of iterations, and attain such a minimum value when approximation error balances noise amplification.
% that both quantities may semi-converge over the greedy iterations, in the sense they reach their respective minimum at a number of iterations smaller than the total number of iterations, and attain such a minimum value when approximation error balances noise amplification.
This is a reflection of the bias--variance tradeoff defined in statistical settings, even if the minima of the two quantities need not coincide (as discussed in the numerical Section \ref{sec:numerical_experiments}).
With reference to Remark~\ref{rmk:bound_trace}, we note that replacing the trace term in SURE with the constant $2 \sigma^2$ (see Equation~\ref{eq:simplificattiontrace}) removes the dependence on the size $j$ of the interpolation set and prevents the typical increase of the tail of the estimator. }
% {This heuristic analogy suggests that the minima of the theoretical error bound and SURE are expected to occur near an iteration that balances approximation error and noise amplification (also referred to as the bias--variance tradeoff in statistical settings).
% Note that replacing the trace term in SURE with the constant $2 \sigma^2$ (see Equation~\ref{eq:simplificattiontrace}) removes the dependence on the size $j$ of the interpolation set and prevents the typical increase of the tail of the estimator. }
% Therefore, the minimum of both theoretical error bound and SURE is reached at the iteration of the greedy method that realizes a tradeoff between approximation error and noise amplification (also called bias-variance tradeoff in statistical settings). 

An example of the tradeoff {and the respective tail behavior} is shown in Figure~\ref{fig:bias_variance_1}, which refers to the numerical experiments subsequently discussed in Section~\ref{sec:numerical_experiments}.

\begin{figure}[!h]
\centering
\includegraphics[width=\textwidth]{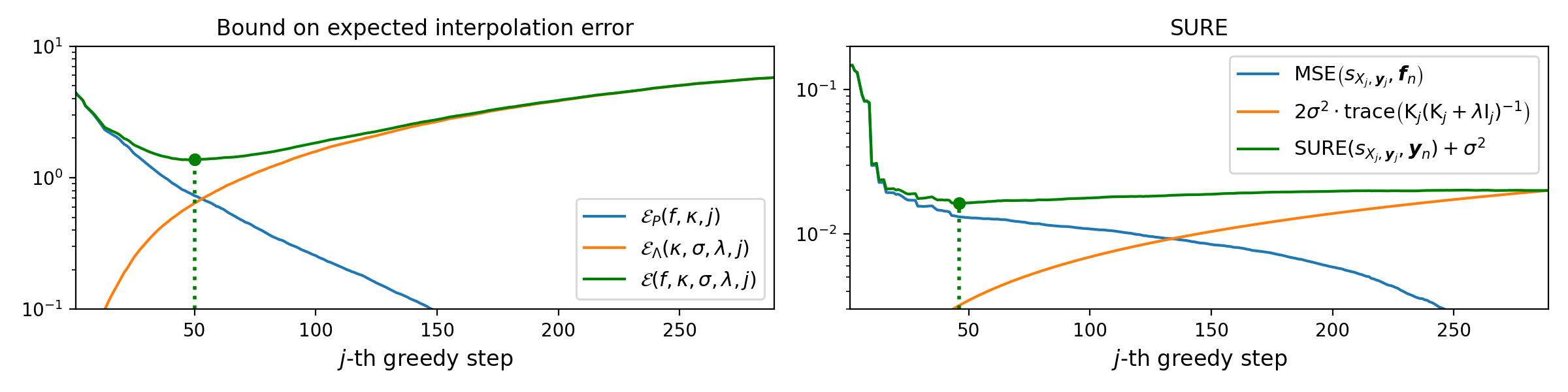}
\caption{Decomposition of $\mathcal{E}(f,\kappa, \sigma, {\lambda}, j)$ and SURE, which are plotted in green in the left and right panel, respectively.
Both quantities are decomposed in a term describing the approximation error (blue line) and a term describing the noise amplification (orange line), as described in Corollary \ref{cor:Expected} and Lemma~\ref{lemma:sure}.
The iterations realizing the minimum of $\mathcal{E}(f,\kappa, \sigma, {\lambda}, j)$ and SURE are indicated with a dashed vertical line.
To avoid plotting negative quantities resulting from the decomposition, $\sigma^2$ is added to value of SURE in the right panel.}
\label{fig:bias_variance}
\end{figure}

\end{remark}

By definition, SURE only depends on the random variable $\boldsymbol{y}_n$ and can hence be computed without knowledge of $\boldsymbol{f}_n$, which is unknown. 
Therefore, in practice, we compute SURE by considering the given realization of the data in place of the random variable.
As we pointed out at the end of Section \ref{sec:interpolation_error}, greedy approaches based on the minimization of $\mathcal{E}(f,\kappa, \sigma, {\lambda}, j)$ are not viable.
Therefore {the proposed a hybrid greedy approach operates as follows}: (1) the nodes are selected independently from the data by minimizing the power function $P^2_{X_j,\kappa}$, and (2) a data-driven {rule based on the minimization of the SURE estimator is employed to identify the greedy iteration that provides the best trade-off between approximation accuracy and noise amplification.}
% rule based on the minimization of the SURE estimator is adopted to determine when to stop the greedy method, preventing noise amplification that occurs once an unnecessarily large number of interpolation nodes is utilized.
The idea is to select the nodes by minimizing only one of the two terms of the theoretical error bound given in \eqref{eq:expected_error_bound_nodes} and to control how the second term increases by monitoring SURE.
{The resulting $p$ESTO procedure is summarized in Algorithm \ref{alg:pESTO}.}

% \begin{algorithm}
%     \KwIn{Nodes $X_n$, observations $\mathbf{y}_n$, kernel $\kappa$, regularization parameter $\lambda$, noise variance $\sigma^2$}

%     Initialize $X_0=\emptyset$
    
%     \For{$j=1,\ldots,n$}{
%         Select the next node by maximizing the power function:
%         \[
%         \mathbf{x}_j=\arg\max_{\mathbf{x}\in X_n\setminus X_{j-1}}
%         P_{X_{j-1},\kappa}(\mathbf{x})
%         \]
        
%         Update $X_j=X_{j-1}\cup\{\mathbf{x}_j\}$
    
%         Compute the regularized interpolant $s_{X_j,\mathbf{y}_j}$
    
%         Evaluate
%         \[
%         \mathrm{SURE}(s_{X_j,\mathbf{y}_j},\mathbf{y}_n)
%         \]
%     }
    
%     Determine
%     \[
%     j^\star=\arg\min_j \mathrm{SURE}(s_{X_j,\mathbf{y}_j},\mathbf{y}_n)
%     \]
    
%     \KwOut{$s_{X_{j^\star},\mathbf{y}_{j^\star}}$}
% \caption{$p$ESTO}\label{alg:pESTO}
% \end{algorithm}
\begin{algorithm}[H]
\small
\caption{\textbf{Pseudo-code of the $p$ESTO algorithm}}
\begin{algorithmic}[1]
%\REQUIRE A collection of nodes
%\ENSURE The greedily selected interpolation set

\item Take an initial interpolation set $X_1=\{\boldsymbol{x}_1\}$. \vskip 0.1cm

\item For $j=1,\ldots,n-1$: \vskip 0.1cm
\begin{enumerate}
    \item[(i)] Define
    $$
    \boldsymbol{x}_{j+1} = \argmax_{\boldsymbol{x} \in X_n\setminus X_j} \,P_{X_j,\kappa}(\boldsymbol{x}) .
    $$
    \vskip 0.1cm

    \item[(ii)] Set
    $$
    X_{j+1}=X_j\cup\{\boldsymbol{x}_{j+1}\}.
    $$
    \vskip 0.1cm

    \item[(iii)] Compute the regularized interpolant
    $s_{X_{j+1},\mathbf{y}_{j+1}}$. \vskip 0.1cm

    \item[(iv)] Compute
    $$
    {\rm SURE}(s_{X_{j+1},\mathbf{y}_{j+1}},\mathbf{y}_n).
    $$
\end{enumerate}

\vskip 0.1cm

\item Select
$$
j^*=
\argmin_{j} \,
{\rm SURE}(s_{X_j,\mathbf{y}_j},\mathbf{y}_n).
$$
\vskip 0.1cm

\item Return $X_{j^*}$. %$s_{X_{j^*},\mathbf{y}_{j^*}}$.

\end{algorithmic}
\label{alg:pESTO}
\end{algorithm}

\begin{remark}
{In the numerical experiments of Section~\ref{sec:numerical_experiments}, we applied the $p$ESTO procedure as presented in Algorithm~\ref{alg:pESTO}.
However, owing to the typical behavior of SURE depicted in Figure~\ref{fig:bias_variance}, in practice, one can stop the execution of the algorithm when the value of SURE plateaus or increases for a given number of iterations. 
% Although Algorithm~\ref{alg:pESTO} evaluates SURE along the entire greedy sequence, in practical implementations the computation may be truncated once the SURE estimate exhibits negligible improvement. In particular, one may introduce a stopping criterion based on the relative variation of SURE between two consecutive iterations, e.g.,
}
% {$$\frac{\left|{\rm SURE}(s_{X_{j+1},\mathbf{y}_{j+1}},\mathbf{y}_n)-{\rm SURE}(s_{X_{j},\mathbf{y}_{j}},\mathbf{y}_n)\right|}
% {\left|{\rm SURE}(s_{X_{j},\mathbf{y}_{j}},\mathbf{y}_n)\right|}<\varepsilon$$}
% {for a prescribed tolerance $\varepsilon$. This may substantially reduce the computational cost when SURE has already reached a plateau, as suggested by the behaviour observed in Figure~\ref{fig:bias_variance}. The criterion is an implementation choice and does not alter the definition of $p$ESTO in Algorithm~\ref{alg:pESTO}.

\end{remark}

% in practice mettiamo un criterio di stop anziché computare tutte le iterazioni (ad esempio quando non miglioro per tot epoche di fila)
% mettiamoci un remark in cui diciamo sta cosa in practice "due to the behaviour observed in figure 1, one could avoid the computation of all the training set choosing/placing a stopping critireon based on the relative impromente betweet the j-th iteration and the j+1-th"

\section{Numerical experiments}\label{sec:numerical_experiments}

{For the numerical experiments, we consider two different kernels that are widely adopted in Radial Basis Function (RBF) theory: the Gaussian kernel and the inverse multiquadric kernel, which are respectively defined as
$$
\kappa_{\rm G}(\boldsymbol{x},\boldsymbol{z}) =\exp\left(-r^2 \left\Vert \boldsymbol{x} - \boldsymbol{z} \right\Vert^2  \right) ~, \quad \kappa_{\rm IMQ}(\boldsymbol{x},\boldsymbol{z}) = 1 / \sqrt{1 + r^2 \left\Vert \boldsymbol{x} - \boldsymbol{z} \right\Vert^2} ~.
$$ 
The shape parameter $r$ is set equal to 10 for both kernels. The optimization of the shape parameter is beyond our scope; however, for details regarding this aspect, we address the reader to, e.g., \cite{Fasshauer15,Ling2022}.}

{
%We consider both a 2D and a 3D interpolation setting.
%As for the former case, 
To assess the effectiveness of the proposed method, we consider two different ground truth functions: the Franke's function \cite{Franke} and the product function \cite{Fasshauer} defined as
$$
f(\boldsymbol{x}) = 16 \, x_1 \, x_2 \, (1-x_1) \, (1-x_2) ~,
$$
where $\boldsymbol{x} = (x_1, x_2)$.
For each kernel and ground truth function, we consider both the Halton interpolation nodes \cite{Halton} and an equispaced grid in the domain $[0,1] \times [0,1]$.
We test the performances of the proposed $p$ESTO using sets of nodes with different cardinality, equal to $13^2$, $17^2$, and $21^2$.
We perturb the data with white Gaussian noise with standard deviation $\sigma$ equal to 0.05 or 0.1, and we generate 30 different realizations for each level of noise.
We perform the proposed $p$-greedy approach from each different data realization. 
In particular, at each step of the greedy method, we construct the regularized interpolant using \eqref{eq:interpolant}, %where we set the value of 
where we empirically set the regularization parameter $\lambda$ equal to $10^{-6}$.
This value of $\lambda$ makes the bound of Remark~\ref{rmk:bound_trace} tight.}

{To quantitatively evaluate the effectiveness of the proposed $p$ESTO approach, we utilize two different metrics: ${\rm MSE}(\mathcal{X}_N)$, which represents the MSE between the regularized interpolant and the ground truth function computed over a fine equispaced grid $\mathcal{X}_N$ of $N = 200 \times 200$ points in $[0,1]\times[0,1]$, and the Structural Similarity Index (SSIM) \cite{ssim_paper}.
Tables~\ref{tab:metrics_franke} and \ref{tab:metrics_valley} report the metric values corresponding to the greedy iteration selected by $p$ESTO.
For comparison, we also report the metric values for the interpolant computer over all the nodes (corresponding to the last greedy iteration, i.e., for $j = n$), and for the optimal greedy iteration.
The latter is defined as the greedy iteration minimizing ${\rm MSE}(\mathcal{X}_N)$, which is, in essence, the target quantity.
For each metric, we report the median and the Median Absolute Deviation (MAD) across the 30 noisy data realizations.
Such indicators are robust to the presence of outliers.}

%%%%%%%%%%%%%%%%%%%%%%%% METRIC FRANKE FUNCTION %%%%%%%%%%%%%%%%%%%%%%%%

\begin{table}[!h]
\centering
\resizebox{\textwidth}{!}{\begin{tabular}{cccccccccc}
\toprule
% \multicolumn{10}{c}{Franke} \\ 
% \midrule
Kernel &Nodes &Num. &$\sigma$  &\multicolumn{3}{c}{${\rm MSE}(\mathcal{X}_N)$} &\multicolumn{3}{c}{SSIM} \\ 
\midrule
& & & &$p$ESTO &All &Opt. &$p$ESTO &All &Opt. \\ 
\midrule

%%%%%%%%% GAUSSIAN HALTON

\multirow{12}{*}[-6ex]{\rotatebox{90}{Gaussian}} &\multirow{6}{*}[-2ex]{Halton} & \multirow{ 2}{*}{$13^2$}  &0.05  &$3.0 \pm $0.8  &$5.2 \pm $0.7  &$2.6 \pm $0.2   &$0.88 \pm $0.04  &$0.76 \pm $0.03  &$0.89 \pm $0.02 \\
\cmidrule{4-10}
&  &  &0.1 &$7.9 \pm $1.8  &$19.8 \pm $2.6  &$7.5 \pm $1.6   &$0.76 \pm $0.06  &$0.49 \pm $0.03  &$0.77 \pm $0.04  \\
\cmidrule{3-10}
&  & \multirow{ 2}{*}{$17^2$}  &0.05 &$2.5 \pm $0.6  &$10.0 \pm $2.4  &$2.3 \pm $0.4   &$0.90 \pm $0.03  &$0.61 \pm $0.04  &$0.90 \pm $0.02  \\
\cmidrule{4-10}
&  &  &0.1 &$7.8 \pm $1.3  &$40.2 \pm $10.0  &$7.5 \pm $1.2   &$0.76 \pm $0.03  &$0.32 \pm $0.04  &$0.77 \pm $0.03  \\
\cmidrule{3-10}
&  & \multirow{ 2}{*}{$21^2$}  &0.05  &$2.0 \pm $0.3  &$27.7 \pm $6.0  &$2.0 \pm $0.2   &$0.91 \pm $0.02  &$0.43 \pm $0.03  &$0.91 \pm $0.02   \\
\cmidrule{4-10}
&  &  &0.1  &$6.8 \pm $0.9  &$110.7 \pm $24.1  &$6.8 \pm $0.8   &$0.78 \pm $0.04  &$0.19 \pm $0.02  &$0.78 \pm $0.04  \\
\cmidrule{2-10}

%%%%%%%%% GAUSSIAN UNIFORM

& \multirow{6}{*}[-2ex]{Equispaced} & \multirow{ 2}{*}{$13^2$}  &0.05  &$2.2 \pm $0.2  &$2.1 \pm $0.3  &$2.1 \pm $0.3   &$0.90 \pm $0.02  &$0.88 \pm $0.01  &$0.89 \pm $0.02  \\
\cmidrule{4-10}
&  &  &0.1  &$8.4 \pm $1.2  &$8.3 \pm $1.1  &$7.7 \pm $1.1   &$0.74 \pm $0.04  &$0.67 \pm $0.02  &$0.75 \pm $0.06  \\
\cmidrule{3-10}
&  & \multirow{ 2}{*}{$17^2$}  &0.05 &$2.2 \pm $0.4  &$2.3 \pm $0.2  &$2.1 \pm $0.2   &$0.91 \pm $0.02  &$0.82 \pm $0.02  &$0.90 \pm $0.03 \\
\cmidrule{4-10}
&  &  &0.1  &$8.0 \pm $1.0  &$9.1 \pm $1.0  &$7.7 \pm $0.8   &$0.76 \pm $0.03  &$0.56 \pm $0.03  &$0.76 \pm $0.03  \\
\cmidrule{3-10}
&  & \multirow{ 2}{*}{$21^2$} &0.05  &$2.3 \pm $0.3  &$2.4 \pm $0.2  &$2.2 \pm $0.2   &$0.88 \pm $0.05  &$0.76 \pm $0.02  &$0.90 \pm $0.02 \\
\cmidrule{4-10}
&   &  &0.1  &$7.3 \pm $1.2  &$9.8 \pm $1.0  &$7.2 \pm $1.1   &$0.77 \pm $0.05  &$0.47 \pm $0.03  &$0.77 \pm $0.05  \\

\midrule

%%%%%%%%%%% INVERSE MULTIQUADRIC - HALTON

\multirow{12}{*}[-6ex]{\rotatebox{90}{Inverse multiquadric}} &\multirow{6}{*}[-2ex]{Halton} & \multirow{ 2}{*}{$13^2$}  &0.05  &$1.7 \pm $0.4  &$2.8 \pm $0.3  &$1.7 \pm $0.3   &$0.94 \pm $0.02  &$0.84 \pm $0.02  &$0.94 \pm $0.02 \\
\cmidrule{4-10}
&  &  &0.1 &$5.8 \pm $1.5  &$11.0 \pm $1.2  &$5.3 \pm $0.9   &$0.86 \pm $0.05  &$0.60 \pm $0.03  &$0.86 \pm $0.04  \\
\cmidrule{3-10}
&  & \multirow{ 2}{*}{$17^2$}  &0.05 &$1.7 \pm $0.3  &$3.3 \pm $0.4  &$1.5 \pm $0.3   &$0.95 \pm $0.02  &$0.76 \pm $0.01  &$0.94 \pm $0.02  \\
\cmidrule{4-10}
&  &  &0.1 &$5.8 \pm $1.7  &$13.3 \pm $1.5  &$5.6 \pm $1.4   &$0.85 \pm $0.04  &$0.49 \pm $0.02  &$0.85 \pm $0.05  \\
\cmidrule{3-10}
&  & \multirow{ 2}{*}{$21^2$}  &0.05  &$1.6 \pm $0.3  &$4.3 \pm $0.3  &$1.6 \pm $0.3   &$0.94 \pm $0.02  &$0.66 \pm $0.01  &$0.94 \pm $0.02   \\
\cmidrule{4-10}
&  &  &0.1  &$5.6 \pm $0.9  &$17.4 \pm $1.3  &$5.4 \pm $0.9   &$0.85 \pm $0.04  &$0.36 \pm $0.02  &$0.84 \pm $0.05  \\
\cmidrule{2-10}

%%%%%%%%%%%%%% INVERSE MULTIQUADRIC - Equispaced

& \multirow{6}{*}[-2ex]{Equispaced} & \multirow{ 2}{*}{$13^2$}  &0.05  &$1.7 \pm $0.4  &$1.8 \pm $0.2  &$1.6 \pm $0.3   &$0.93 \pm $0.02  &$0.90 \pm $0.01  &$0.94 \pm $0.03  \\
\cmidrule{4-10}
&  &  &0.1  &$6.0 \pm $1.4  &$7.4 \pm $1.0  &$5.3 \pm $1.0   &$0.84 \pm $0.05  &$0.70 \pm $0.02  &$0.84 \pm $0.05  \\
\cmidrule{3-10}
&  & \multirow{ 2}{*}{$17^2$}  &0.05 &$1.6 \pm $0.2  &$2.0 \pm $0.2  &$1.5 \pm $0.3   &$0.94 \pm $0.02  &$0.84 \pm $0.02  &$0.94 \pm $0.02 \\
\cmidrule{4-10}
&  &  &0.1  &$4.7 \pm $1.0  &$8.0 \pm $0.7  &$4.6 \pm $1.1   &$0.89 \pm $0.04  &$0.60 \pm $0.02  &$0.89 \pm $0.05 \\
\cmidrule{3-10}
&  & \multirow{ 2}{*}{$21^2$} &0.05  &$1.7 \pm $0.3  &$2.1 \pm $0.1  &$1.6 \pm $0.3   &$0.93 \pm $0.02  &$0.79 \pm $0.02  &$0.94 \pm $0.02 \\
\cmidrule{4-10}
&   &  &0.1  &$5.7 \pm $1.5  &$8.4 \pm $0.6  &$5.5 \pm $1.3   &$0.85 \pm $0.06  &$0.52 \pm $0.03  &$0.84 \pm $0.06 \\

\bottomrule
\end{tabular}}
\caption{{Median and MAD of the ${\rm MSE}(\mathcal{X}_N)$ and SSIM metrics for the ground truth Franke's function computed across 30 noisy data realizations.
% For each kernel, type of interpolation nodes, cardinality of the interpolation nodes, and level of noise, we report the metric value of the greedy iteration selected by $p$ESTO, of the last greedy iteration (all nodes), and of the optimal greedy iteration. 
% The latter is defined as the greedy iteration at which ${\rm MSE}(\mathcal{X}_N)$ attains the minimum value.
}}
\label{tab:metrics_franke}
\end{table}

% %%%%%%%%%%%%%%%%%%%%%%%% METRIC FRANKE PRODUCT %%%%%%%%%%%%%%%%%%%%%%%%

\begin{table}[!h]
\centering
\resizebox{\textwidth}{!}{\begin{tabular}{cccccccccc}
\toprule
% \multicolumn{10}{c}{Franke} \\ 
% \midrule
Kernel &Nodes &Num. &$\sigma$  &\multicolumn{3}{c}{${\rm MSE}(\mathcal{X}_N)$} &\multicolumn{3}{c}{SSIM} \\ 
\midrule
& & & &$p$ESTO &All &Opt. &$p$ESTO &All &Opt. \\ 
\midrule

%%%%%%%%%%%%%%%%% GAUSSIAN - HALTON

\multirow{12}{*}[-6ex]{\rotatebox{90}{Gaussian}} &\multirow{6}{*}[-2ex]{Halton} & \multirow{ 2}{*}{$13^2$}  &0.05 &$3.2 \pm $0.9  &$5.1 \pm $0.8  &$2.8 \pm $0.4   &$0.85 \pm $0.05  &$0.75 \pm $0.03  &$0.88 \pm $0.01 \\
\cmidrule{4-10}
&  &  &0.1 &$9.5 \pm $2.6  &$20.0 \pm $3.1  &$8.2 \pm $1.4   &$0.70 \pm $0.07  &$0.48 \pm $0.03  &$0.73 \pm $0.04  \\
\cmidrule{3-10}
&  & \multirow{ 2}{*}{$17^2$}  &0.05 &$2.6 \pm $0.6  &$10.2 \pm $2.6  &$2.4 \pm $0.4   &$0.87 \pm $0.04  &$0.59 \pm $0.04  &$0.88 \pm $0.02  \\
\cmidrule{4-10}
&  &  &0.1  &$8.3 \pm $1.8  &$40.7 \pm $10.4  &$7.9 \pm $1.7   &$0.74 \pm $0.04  &$0.31 \pm $0.04  &$0.75 \pm $0.03 \\
\cmidrule{3-10}
&  & \multirow{ 2}{*}{$21^2$}  &0.05  &$2.3 \pm $0.4  &$27.7 \pm $6.1  &$2.2 \pm $0.3   &$0.88 \pm $0.02  &$0.42 \pm $0.03  &$0.88 \pm $0.02  \\
\cmidrule{4-10}
&  &  &0.1  &$7.5 \pm $1.1  &$110.7 \pm $24.4  &$7.4 \pm $1.2   &$0.75 \pm $0.03  &$0.18 \pm $0.02  &$0.76 \pm $0.03  \\
\cmidrule{2-10}

%%%%%%%%%%%%%%%%% GAUSSIAN - Equispaced

& \multirow{6}{*}[-2ex]{Equispaced} & \multirow{ 2}{*}{$13^2$}  &0.05 &$2.3 \pm $0.2  &$2.1 \pm $0.3  &$2.1 \pm $0.3   &$0.87 \pm $0.02  &$0.87 \pm $0.01  &$0.87 \pm $0.02 \\
\cmidrule{4-10}
&  &  &0.1  &$8.5 \pm $1.8  &$8.3 \pm $1.0  &$8.1 \pm $1.1   &$0.71 \pm $0.05  &$0.66 \pm $0.02  &$0.69 \pm $0.05 \\
\cmidrule{3-10}
&  & \multirow{ 2}{*}{$17^2$}  &0.05  &$2.3 \pm $0.2  &$2.3 \pm $0.2  &$2.2 \pm $0.2   &$0.89 \pm $0.03  &$0.81 \pm $0.02  &$0.88 \pm $0.05 \\
\cmidrule{4-10}
&  &  &0.1 &$8.0 \pm $0.9  &$9.2 \pm $1.0  &$7.8 \pm $0.7   &$0.75 \pm $0.03  &$0.55 \pm $0.03  &$0.75 \pm $0.03 \\
\cmidrule{3-10}
&  & \multirow{ 2}{*}{$21^2$} &0.05  &$2.3 \pm $0.4  &$2.4 \pm $0.2  &$2.2 \pm $0.3   &$0.87 \pm $0.03  &$0.76 \pm $0.02  &$0.87 \pm $0.03 \\
\cmidrule{4-10}
&   &  &0.1 &$7.8 \pm $1.4  &$9.8 \pm $0.9  &$7.7 \pm $1.1   &$0.74 \pm $0.04  &$0.46 \pm $0.03  &$0.75 \pm $0.04 \\

%%%%%%%%%%%%%%%%% INVERSE MULTIQUADRIC - HALTON

\midrule

\multirow{12}{*}[-6ex]{\rotatebox{90}{Inverse multiquadric}} &\multirow{6}{*}[-2ex]{Halton} & \multirow{ 2}{*}{$13^2$}  &0.05 &$2.2 \pm $0.4  &$3.0 \pm $0.3  &$2.0 \pm $0.2   &$0.91 \pm $0.01  &$0.83 \pm $0.02  &$0.91 \pm $0.01 \\
\cmidrule{4-10}
&  &  &0.1  &$6.7 \pm $1.4  &$11.2 \pm $1.2  &$6.2 \pm $1.0   &$0.83 \pm $0.03  &$0.60 \pm $0.03  &$0.84 \pm $0.03 \\
\cmidrule{3-10}
&  & \multirow{ 2}{*}{$17^2$}  &0.05  &$1.8 \pm $0.3  &$3.3 \pm $0.3  &$1.7 \pm $0.2   &$0.92 \pm $0.02  &$0.75 \pm $0.02  &$0.92 \pm $0.02 \\
\cmidrule{4-10}
&  &  &0.1  &$6.1 \pm $1.3  &$13.2 \pm $1.3  &$5.9 \pm $1.2   &$0.84 \pm $0.04  &$0.46 \pm $0.02  &$0.84 \pm $0.04 \\
\cmidrule{3-10}
&  & \multirow{ 2}{*}{$21^2$}  &0.05 &$1.7 \pm $0.3  &$4.4 \pm $0.3  &$1.6 \pm $0.3   &$0.92 \pm $0.02  &$0.65 \pm $0.01  &$0.92 \pm $0.01 \\
\cmidrule{4-10}
&  &  &0.1 &$5.6 \pm $1.1  &$17.4 \pm $1.3  &$5.5 \pm $1.2   &$0.84 \pm $0.03  &$0.34 \pm $0.01  &$0.83 \pm $0.04 \\
\cmidrule{2-10}

%%%%%%%%%%%%%%%%% INVERSE MULTIQUADRIC - Equispaced

& \multirow{6}{*}[-2ex]{Equispaced} & \multirow{ 2}{*}{$13^2$}  &0.05  &$1.9 \pm $0.4  &$1.9 \pm $0.2  &$1.7 \pm $0.2   &$0.91 \pm $0.01  &$0.88 \pm $0.01  &$0.92 \pm $0.02 \\
\cmidrule{4-10}
&  &  &0.1  &$6.3 \pm $1.1  &$7.4 \pm $1.0  &$5.8 \pm $0.8   &$0.82 \pm $0.05  &$0.69 \pm $0.02  &$0.84 \pm $0.03 \\
\cmidrule{3-10}
&  & \multirow{ 2}{*}{$17^2$}  &0.05  &$1.7 \pm $0.3  &$2.0 \pm $0.2  &$1.6 \pm $0.3   &$0.92 \pm $0.02  &$0.84 \pm $0.01  &$0.92 \pm $0.02 \\
\cmidrule{4-10}
&  &  &0.1  &$5.9 \pm $1.4  &$8.1 \pm $0.8  &$5.4 \pm $1.6   &$0.84 \pm $0.05  &$0.59 \pm $0.03  &$0.85 \pm $0.04 \\
\cmidrule{3-10}
&  & \multirow{ 2}{*}{$21^2$} &0.05  &$1.7 \pm $0.4  &$2.1 \pm $0.1  &$1.7 \pm $0.4   &$0.92 \pm $0.02  &$0.79 \pm $0.01  &$0.92 \pm $0.01 \\
\cmidrule{4-10}
&   &  &0.1  &$6.0 \pm $1.4  &$8.4 \pm $0.6  &$5.8 \pm $1.2   &$0.85 \pm $0.04  &$0.51 \pm $0.02  &$0.83 \pm $0.04 \\

\bottomrule
\end{tabular}}
\caption{{Median and MAD of the ${\rm MSE}(\mathcal{X}_N)$ and SSIM metrics for the ground truth product function computed across 30 noisy data realizations.}}
\label{tab:metrics_valley}
\end{table}

{Results shown in Tables~\ref{tab:metrics_franke} and \ref{tab:metrics_valley} demonstrate that the subsets of nodes selected by $p$ESTO lead to performances comparable to those of the optimal subset of nodes.
In fact, the median metric values corresponding to $p$ESTO are systematically consistent within MAD with those corresponding to the optimal subset of nodes.
Further, the $p$ESTO stopping criterion almost systematically yields improved performances compared to the use of all nodes. 
The only exceptions occur in the case of interpolation nodes located on an equispaced grid, where sometimes the median ${\rm MSE}(\mathcal{X}_N)$ values for the interpolant obtained with all nodes are slighly smaller (although comparable within MAD).
However, for most of these exceptions, the SSIM metric of $p$ESTO is larger than that corresponding to all nodes, showing the $p$ESTO leads to comparable performances using a smaller number of nodes.}

{Table~\ref{tab:greedy_iteration} compares the greedy iterations selected by the proposed $p$ESTO with the optimal ones (i.e., those minimizing the ${\rm MSE}(\mathcal{X}_N)$ metric).
Specifically, we report the median and MAD of the selected iterations computed across the 30 noise realizations. 
Results show that the median of the greedy iterations selected by $p$ESTO is consistent within MAD with the median of the optimal ones.
Therefore, $p$ESTO represents a good proxy for ${\rm MSE}(\mathcal{X}_N)$, although it does not make use of evaluations of the ground truth function.
Only in three cases corresponding to the Gaussian kernel and $13^2$ equispaced interpolation nodes, the optimal iteration appears to be close to the last one. 
In this cases, $p$ESTO selects a sensibly smaller number of nodes, however, the corresponding metrics shown in Tables~\ref{tab:metrics_franke} and \ref{tab:metrics_valley} demonstrate that the performances of $p$ESTO are comparable to the optimal ones with fewer interpolation nodes.
}

\begin{table}[!h]
\centering
\resizebox{\textwidth}{!}{
% \begin{tabular}{c|c|c|c|c|c|c|c}
\begin{tabular}{cccccccc}
\toprule
Kernel &Nodes &Num. &$\sigma$  &\multicolumn{4}{c}{Selected greedy iteration}  \\ 

\midrule
& & & &\multicolumn{2}{c}{Franke} &\multicolumn{2}{c}{Product function} \\ 
\midrule

& & & &$p$ESTO &Opt. &$p$ESTO &Opt. \\ 
\midrule

\multirow{12}{*}[-6ex]{\rotatebox{90}{Gaussian}} &\multirow{6}{*}[-2ex]{Halton} & \multirow{ 2}{*}{$13^2$} &0.05 &$100.0 \pm $25.9  &$102.0 \pm $10.4 &$108.0 \pm $35.6  &$96.5 \pm $20.0 \\
\cmidrule{4-8}
&  &  &0.1 &$64.5 \pm $9.6  &$69.0 \pm $7.4 &$84.5 \pm $21.5  &$80.0 \pm $9.6 \\
\cmidrule{3-8}
&  &\multirow{ 2}{*}{$17^2$}  &0.05 &$98.5 \pm $27.4  &$88.5 \pm $12.6 &$123.0 \pm $34.1  &$111.5 \pm $21.5 \\
\cmidrule{4-8}
&  &  &0.1 &$74.5 \pm $9.6  &$76.0 \pm $10.4 &$84.5 \pm $13.3  &$79.0 \pm $8.2 \\
\cmidrule{3-8}
&  &\multirow{ 2}{*}{$21^2$}  &0.05 &$91.0 \pm $24.5  &$91.0 \pm $21.5 &$123.5 \pm $41.5  &$125.5 \pm $26.7 \\
\cmidrule{4-8}
&  &  &0.1 &$73.5 \pm $3.0  &$74.5 \pm $5.2 &$81.5 \pm $15.6  &$81.5 \pm $15.6 \\

\cmidrule{2-8}

& \multirow{6}{*}[-2ex]{Equispaced} & \multirow{ 2}{*}{$13^2$} &0.05 &$119.0 \pm $29.7  &$167.0 \pm $3.0 &$114.0 \pm $23.7  &$168.5 \pm $0.7 \\
\cmidrule{4-8}
&  &  &0.1 &$99.0 \pm $11.9  &$97.5 \pm $8.9 &$99.5 \pm $10.4  &$167.0 \pm $3.0 \\
\cmidrule{3-8}
&  &\multirow{ 2}{*}{$17^2$}  &0.05 &$93.0 \pm $8.9  &$103.0 \pm $21.5 &$98.0 \pm $18.5  &$115.5 \pm $35.6 \\
\cmidrule{4-8}
&  &  &0.1 &$86.5 \pm $8.2  &$87.0 \pm $9.6 &$80.0 \pm $2.2  &$81.5 \pm $5.2 \\
\cmidrule{3-8}
&  &\multirow{ 2}{*}{$21^2$}  &0.05 &$111.0 \pm $22.2  &$114.0 \pm $15.6 &$129.5 \pm $28.2  &$125.5 \pm $31.9 \\
\cmidrule{4-8}
&  &  &0.1 &$74.0 \pm $5.2  &$71.0 \pm $5.9 &$72.5 \pm $8.9  &$83.0 \pm $20.0 \\

\midrule

\multirow{12}{*}[-6ex]{\rotatebox{90}{Inverse multiquadric}} &\multirow{6}{*}[-2ex]{Halton} & \multirow{ 2}{*}{$13^2$} &0.05  &$49.5 \pm $19.3  &$56.0 \pm $18.5 &$63.0 \pm $19.3  &$75.0 \pm $23.0 \\
\cmidrule{4-8}
&  &  &0.1 &$35.0 \pm $8.9  &$35.0 \pm $9.6 &$41.0 \pm $8.2  &$38.0 \pm $7.4 \\
\cmidrule{3-8}
&  &\multirow{ 2}{*}{$17^2$}  &0.05 &$53.5 \pm $14.8  &$72.0 \pm $29.7 &$64.0 \pm $18.5  &$78.0 \pm $28.9 \\
\cmidrule{4-8}
&  &  &0.1 &$44.5 \pm $4.4  &$49.0 \pm $17.0 &$45.5 \pm $11.9  &$52.0 \pm $23.0 \\
\cmidrule{3-8}
&  &\multirow{ 2}{*}{$21^2$}  &0.05 &$77.0 \pm $24.5  &$77.0 \pm $26.7 &$69.5 \pm $24.5  &$69.0 \pm $24.5 \\
\cmidrule{4-8}
&  &  &0.1 &$48.0 \pm $14.1  &$48.5 \pm $15.6 &$48.0 \pm $14.8  &$48.0 \pm $14.8 \\

\cmidrule{2-8}

& \multirow{6}{*}[-2ex]{Equispaced} & \multirow{ 2}{*}{$13^2$} &0.05 &$54.0 \pm $8.2  &$80.5 \pm $40.8 &$74.5 \pm $17.8  &$97.5 \pm $60.8 \\
\cmidrule{4-8}
&  &  &0.1 &$48.0 \pm $18.5  &$50.0 \pm $18.5 &$44.0 \pm $17.0  &$46.0 \pm $14.1 \\
\cmidrule{3-8}
&  &\multirow{ 2}{*}{$17^2$}  &0.05 &$72.0 \pm $22.2  &$79.0 \pm $13.3 &$79.0 \pm $22.2  &$79.0 \pm $25.2 \\
\cmidrule{4-8}
&  &  &0.1 &$38.0 \pm $17.8  &$39.0 \pm $19.3 &$51.0 \pm $17.8  &$59.5 \pm $24.5 \\
\cmidrule{3-8}
&  &\multirow{ 2}{*}{$21^2$}  &0.05 &$80.0 \pm $31.1  &$83.0 \pm $27.4 &$92.0 \pm $35.6  &$76.0 \pm $28.2 \\
\cmidrule{4-8}
&  &  &0.1 &$42.0 \pm $21.5  &$42.5 \pm $23.0 &$48.5 \pm $15.6  &$53.5 \pm $23.0 \\

\bottomrule
\end{tabular}
}
\caption{Median and MAD of the greedy iterations selected by $p$ESTO and of the optimal greedy iterations computed across 30 noisy data realizations.
% (i.e., those at which ${\rm MSE}(\mathcal{X}_N)$ attains its minimum value).
% For each kernel, type of interpolation nodes, cardinality of the interpolation nodes, level of noise, and ground truth function, the median and MAD are computed across 30 noisy realizations of the data. 
}
\label{tab:greedy_iteration}
\end{table}

\begin{figure}[!h]
\centering
\includegraphics[width=\textwidth]{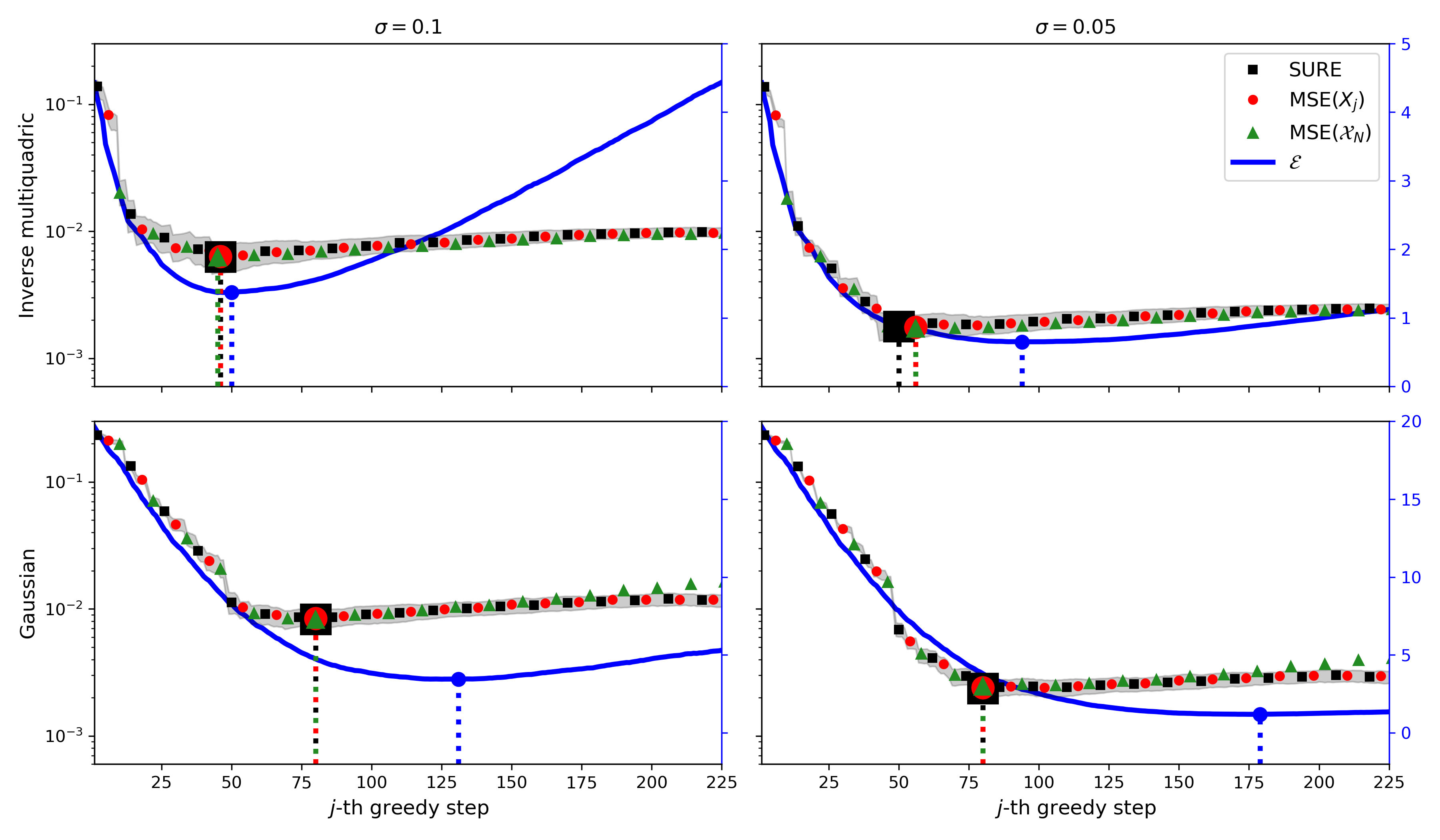}
\caption{{Results for the ground truth Franke's function and $17^2$ Halton interpolation nodes.
The dashed vertical lines indicate the greedy iteration at which the corresponding quantities attain their minimum value.
Dependencies in $\mathcal{E}$ are omitted and, for short, ${\rm MSE}(X_j)$ indicates ${\rm MSE}(s_{X_j, \boldsymbol{y}_j}, \boldsymbol{f}_n)$.
}
}
\label{fig:results_franke}
\end{figure}

\begin{figure}[!h]
\centering
\includegraphics[width=\textwidth]{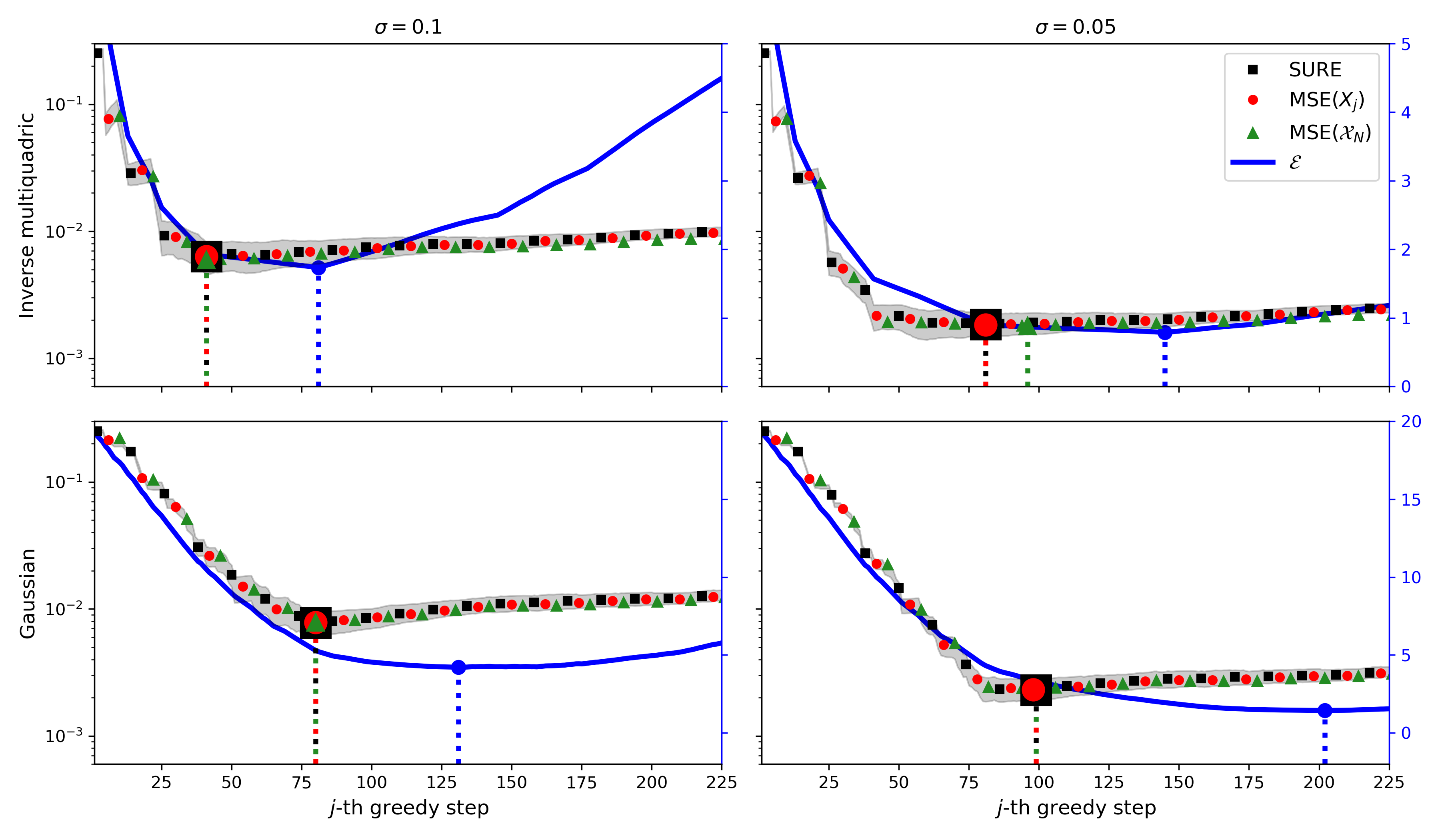}
\caption{{Results for the ground truth product function and $17^2$ interpolation nodes on an equispaced grid.
The dashed vertical lines indicate the greedy iteration at which the corresponding quantities attain their minimum value.
Dependencies in $\mathcal{E}$ are omitted and, for short, ${\rm MSE}(X_j)$ indicates ${\rm MSE}(s_{X_j, \boldsymbol{y}_j}, \boldsymbol{f}_n)$.}
}
\label{fig:results_product}
\end{figure}

{To visually appreciate the behaviour of SURE and compare it with the theoretical error bound derived in Section~\ref{sec:interpolation_error}, in Figures~\ref{fig:results_franke} and \ref{fig:results_product} we display the sample mean of SURE, ${\rm MSE}(s_{X_j, \boldsymbol{y}_j}, \boldsymbol{f}_n)$, and ${\rm MSE}(\mathcal{X}_N)$ computed across the 30 noisy realizations, for the Franke's and the product function, respectively.
The shaded area corresponds to a region within one standard deviation around the mean of SURE.
In all cases, the sample mean of ${\rm MSE}(s_{X_j, \boldsymbol{y}_j}, \boldsymbol{f}_n)$ falls within the shaded area, thus showing that SURE is an unbiased estimator of ${\rm MSE}(s_{X_j, \boldsymbol{y}_j}, \boldsymbol{f}_n)$.
Further, consistent with the results presented in Table~\ref{tab:greedy_iteration}, the minimum of the sample mean of ${\rm MSE}(\mathcal{X}_N)$ is close to the SURE one.
It is interesting to observe that the theoretical error bound attains a minimum value at a generally larger greedy iteration.
This suggests that considering the bound on the expected interpolation error as a stopping rule leads to a less computationally efficient strategy even assuming that $\Vert f \Vert_{\mathcal{H}_\kappa}$ is known.
}

\section{Conclusion}
\label{Conclusion}

% We proposed $p$ESTO, a risk-driven stopping rule for iterative $p$-greedy kernel regression in the presence of noisy data. The approach relies on probabilistic error bounds that separate approximation accuracy and noise amplification, and leverages SURE to estimate the MSE without access to the true solution. Numerical results confirm that $p$ESTO effectively identifies the \emph{optimal} stopping rule. Future work will focus on extending the proposed framework to non-Gaussian noise models, in particular to Poisson noise.

In this work, we analyzed error bounds for kernel approximation in the presence of noisy data.
In such a setting, usual error bounds based on the power function must be corrected by a statistical term.
The balance of the two contributions is unknown, since the norm of the exact function is not available; as a consequence, bounds that separate approximation accuracy and noise amplification turn out to be ineffective.
We proposed $p$ESTO, a risk-driven stopping rule for iterative $p$-greedy kernel regression in the presence of noisy data, which leverages SURE to estimate the MSE without access to the true solution. Numerical results confirm that $p$ESTO effectively identifies the \emph{optimal} greedy iteration. 
Future work will focus on extending the proposed framework to non-Gaussian noise models, in particular to Poisson noise \cite{massa2021predictive}.

\section*{Acknowledgements}
MT and EP acknowledge the support of the Fondazione Compagnia di San Paolo within the framework of the Artificial Intelligence Call for Proposals, AIxtreme project (ID Rol: 71708). LBB and EP are further supported by the GOSSIP project (``Greedy Optimal Sampling for Solar Inverse Problems'') – funded by the Ministero dell’Universit\`a e della Ricerca (MUR) -  within the PRIN 2022 program, CUP: E53C24002330001. LBB is also funded by the IN$\delta$AM-GNCS project CUP: E53C25002010001.
PM is supported by Swiss PRODEX grant for STIX.

\bibliographystyle{plain}

\begin{thebibliography}{10}
\expandafter\ifx\csname url\endcsname\relax
  \def\url#1{\texttt{#1}}\fi
\expandafter\ifx\csname urlprefix\endcsname\relax\def\urlprefix{URL }\fi
\expandafter\ifx\csname href\endcsname\relax
  \def\href#1#2{#2} \def\path#1{#1}\fi

\bibitem{Fasshauer15}
G.~E. Fasshauer, M.~J. McCourt, Kernel-based Approximation Methods using {MATLAB}, World scientific, 2015.

\bibitem{santin2018greedyregularizedkernelinterpolation}
G.~Santin, D.~Wittwar, B.~Haasdonk, Greedy regularized kernel interpolation (2018).
\newblock \href {http://arxiv.org/abs/1807.09575} {\path{arXiv:1807.09575}}.

\bibitem{WenzelSantinHaasdonk}
T.~Wenzel, G.~Santin, B.~Haasdonk, Analysis of target data-dependent greedy kernel algorithms: convergence rates for {$f$}-, {$f\cdot P$}-, and {$f/P$}-greedy, Constr. Approx. 57~(1) (2023) 45--74.

\bibitem{morozov1966regularization}
V.~A. Morozov, Regularization of incorrectly posed problems and the choice of regularization parameter, USSR Computational Mathematics and Mathematical Physics 6~(1) (1966) 242--251.

\bibitem{GCV}
G.~H. Golub, M.~Heath, G.~Wahba, Generalized cross-validation as a method for choosing a good ridge parameter, Technometrics 21~(2) (1979) 215--223.

\bibitem{BENVENUTO20191}
F.~Benvenuto, C.~Campi, A discrepancy principle for the landweber iteration based on risk minimization, Applied Mathematics Letters 96 (2019) 1--6.

\bibitem{stein1981estimation}
C.~M. Stein, Estimation of the mean of a multivariate normal distribution, The annals of Statistics (1981) 1135--1151.

\bibitem{raskutti2014early}
G.~Raskutti, M.~J. Wainwright, B.~Yu, Early stopping and non-parametric regression: an optimal data-dependent stopping rule, The Journal of Machine Learning Research 15~(1) (2014) 335--366.

\bibitem{bruno2025greedytechniquesinverseproblems}
L.~{Bruni Bruno}, P.~Massa, E.~Perracchione, M.~Trombini, Greedy techniques for inverse problems (2025).
\newblock \href {http://arxiv.org/abs/2512.04046} {\path{arXiv:2512.04046}}.

\bibitem{De_Marchi_Marchetti_Perracchione_2026}
S.~De~Marchi, F.~Marchetti, E.~Perracchione, Kernel-Based Methods: Milestones and New Trends, Cambridge Monographs on Applied and Computational Mathematics, Cambridge University Press, 2026.

\bibitem{Wendland_2004}
H.~Wendland, Scattered Data Approximation, Cambridge Monographs on Applied and Computational Mathematics, Cambridge University Press, 2004.

\bibitem{GenReprThm}
B.~Sch\"{o}lkopf, R.~Herbrich, A.~J. Smola, A generalized representer theorem, in: Computational learning theory ({A}msterdam, 2001), Vol. 2111 of Lecture Notes in Comput. Sci., Springer, Berlin, 2001, pp. 416--426.

\bibitem{DeMarchi2005317}
S.~De~Marchi, R.~Schaback, H.~Wendland, Near-optimal data-independent point locations for radial basis function interpolation, Advances in Computational Mathematics 23~(3) (2005) 317--330.

\bibitem{Wenzel2021}
T.~Wenzel, G.~Santin, B.~Haasdonk, A novel class of stabilized greedy kernel approximation algorithms: Convergence, stability and uniform point distribution, Journal of Approximation Theory 262 (2021).

\bibitem{Themistoclakis}
W.~Themistoclakis, M.~Van~Barel, A new kernel method for the uniform approximation in reproducing kernel {H}ilbert spaces, Appl. Math. Lett. 153 (2024) Paper No. 109052, 8.

\bibitem{kim2024enhancinggaussianprocesssurrogates}
H.~Kim, D.~Sanz-Alonso, Enhancing gaussian process surrogates for optimization and posterior approximation via random exploration (2024).
\newblock \href {http://arxiv.org/abs/2401.17037} {\path{arXiv:2401.17037}}.

\bibitem{wenzel2026reliablesamplingbasedrkhsnorm}
T.~Wenzel, A.~Tokmak, C.~Fiedler, Reliable sampling-based rkhs norm estimation via superconvergence (2026).
\newblock \href {http://arxiv.org/abs/2605.20091} {\path{arXiv:2605.20091}}.

\bibitem{MR2208349}
F.~M. Dekking, C.~Kraaikamp, H.~P. Lopuha\"{a}, L.~E. Meester, A modern introduction to probability and statistics, Springer Texts in Statistics, Springer-Verlag London, Ltd., London, 2005.

\bibitem{Benvenuto_2020}
F.~Benvenuto, B.~Jin, A parameter choice rule for tikhonov regularization based on predictive risk, Inverse Problems 36~(6) (2020) 065004.

\bibitem{Deledalle_2017}
C.-A. Deledalle, Estimation of {K}ullback-{L}eibler losses for noisy recovery problems within the exponential family, Electron. J. Stat. 11~(2) (2017) 3141--3164.

\bibitem{Ling2022}
L.~Ling, F.~Marchetti, A stochastic extended {R}ippa's algorithm for {LpOCV}, Applied Mathematics Letters 129 (2022).

\bibitem{Franke}
R.~Franke, Scattered data interpolation: tests of some methods, Math. Comp. 38~(157) (1982) 181--200.

\bibitem{Fasshauer}
G.~E. Fasshauer, Meshfree approximation methods with {MATLAB}, Vol.~6 of Interdisciplinary Mathematical Sciences, World Scientific Publishing Co. Pte. Ltd., Hackensack, NJ, 2007.

\bibitem{Halton}
J.~H. Halton, On the efficiency of certain quasi-random sequences of points in evaluating multi-dimensional integrals, Numer. Math. 2 (1960) 84--90.

\bibitem{ssim_paper}
Z.~Wang, A.~Bovik, H.~Sheikh, E.~Simoncelli, Image quality assessment: from error visibility to structural similarity, IEEE Transactions on Image Processing 13~(4) (2004) 600--612.

\bibitem{massa2021predictive}
P.~Massa, F.~Benvenuto, Predictive risk estimation for the expectation maximization algorithm with poisson data, Inverse Problems 37~(4) (2021) 045013.

\end{thebibliography}

\end{document}